\documentclass[
]{ceurart}

\usepackage{listings}

\usepackage{algorithm}
\usepackage{algpseudocode}
\usepackage{algorithmicx}
\usepackage{tcolorbox}
\begin{document}

\copyrightyear{2025}
\copyrightclause{Copyright for this paper by its authors.
  Use permitted under Creative Commons License Attribution 4.0
  International (CC BY 4.0).}

% \conference{Woodstock'22: Symposium on the irreproducible science,
%   June 07--11, 2022, Woodstock, NY}

\title{Towards an Automated Multimodal Approach for Video Summarization: Building a Bridge Between Text, Audio and Facial Cue-Based Summarization}

% \tnotetext[1]{You can use this document as the template for preparing your
%   publication. We recommend using the latest version of the ceurart style.}

\author[1]{Md Moinul Islam}[%
% orcid=0000-0002-0877-7063,
email=moinul.islam@student.oulu.fi,
% url=https://yamadharma.github.io/,
]
\cormark[1]
\address[1]{Center for Machine Vision and Signal Analysis, Faculty of ITEE, University of Oulu, Finland}
% \address[2]{Joint Institute for Nuclear Research,
%   6 Joliot-Curie, Dubna, Moscow region, 141980, Russian Federation}
\author[1]{Sofoklis Kakouros}

\author[1]{Janne Heikkilä}

\author[1]{Mourad Oussalah}[%
% orcid=0000-0001-7116-9338,
email=mourad.oussalah@oulu.fi,
% url=https://kmitd.github.io/ilaria/,
]
% \address[1]{Centre for Machine Vision and Signal Analysis),
%   University of Oulu, Finland}

% \author[4]{Manfred Jeusfeld}[%
% orcid=0000-0002-9421-8566,
% email=Manfred.Jeusfeld@acm.org,
% url=http://conceptbase.sourceforge.net/mjf/,
% \address[4]{University of Skövde, Högskolevägen 1, 541 28 Skövde, Sweden}

\cortext[1]{Corresponding author.}
% \fntext[1]{These authors contributed equally.}


\begin{abstract}
The increasing volume of video content in educational, professional, and social domains necessitates effective summarization techniques that go beyond traditional unimodal approaches. This paper proposes a behaviour-aware multimodal video summarization framework that integrates textual, audio, and visual cues to generate timestamp-aligned summaries. By extracting prosodic features, textual cues and visual indicators, the framework identifies semantically and emotionally important moments. A key contribution is the identification of \textit{bonus words}, which are terms emphasized across multiple modalities and used to improve the semantic relevance and expressive clarity of the summaries. The approach is evaluated against pseudo-ground truth (pGT) summaries generated using LLM-based extractive method. Experimental results demonstrate significant improvements over traditional extractive method, such as the Edmundson method, in both text and video-based evaluation metrics. Text-based metrics show ROUGE-1 increasing from 0.4769 to 0.7929 and BERTScore from 0.9152 to 0.9536, while in video-based evaluation, our proposed framework improves F1-Score by almost 23\%. The findings underscore the potential of multimodal integration in producing comprehensive and behaviourally informed video summaries.
\end{abstract}


\conference{HHAI-WS 2025: Workshops at the Fourth International Conference on Hybrid Human-Artificial Intelligence (HHAI), June 9–13, 2025, Pisa, Italy}
\maketitle

\section{Introduction}
\label{sec:1}

In recent years, the consumption of video content has reached unprecedented levels, becoming a central part of modern communication, education, and professional interactions. Platforms hosting interviews, lectures, and virtual meetings have become essential tools for sharing knowledge and fostering collaboration. Unlike static forms of media such as text or images, video integrates spoken language, facial expressions, and vocal tone, providing a multi-sensory experience that enhances the richness of the conveyed information. However, this abundance of video content brings with it a challenge: information overload. Viewers often face the difficult task of navigating through long videos to extract key insights, a process that becomes increasingly impractical as the volume of content continues to grow. This has led to the development of automatic video summarization systems, which aim to condense long videos into concise summaries that retain the core messages and important context \cite{apostolidis2021video}.

Traditional video summarization methods have been largely based on unimodal techniques. Visual-based methods, such as those that detect scene boundaries or extract key-frames, are effective in analyzing the structural aspects of the video. However, these approaches often miss the deeper meaning conveyed through audio or behavioral cues \cite{chu2015video}. On the other hand, text-based methods, which summarize transcribed speech, capture linguistic content but fail to consider the emotional and contextual nuances provided by visual and auditory features, such as tone of voice or facial expressions \cite{tiwari2021survey}. These single modality-based approaches struggle to fully represent the richness of human communication, where meaning often arises from the integration of multiple sensory elements \cite{otani2017video}.

To address these limitations, multimodal video summarization has emerged as a promising solution. By combining visual, auditory, and textual cues, multimodal methods can provide a more comprehensive understanding of content. Visual cues, such as facial expressions and body posture, provide insight into the speaker’s emotions and communicative intent, while auditory features, such as pitch and loudness, highlight emotional states. Textual data, aligned with the video timeline, provides semantic context, enabling a more accurate and contextually rich summary \cite{rochan2018video}.

In this work, we propose a novel multimodal video summarization framework that integrates audio, visual, and textual cues to identify important video segments. The framework focuses on identifying significant changes in behavior, such as variations in vocal intensity or notable gestures, and correlating these with spoken content to produce more relevant summaries. One key feature of our approach is the identification of \textit{bonus words}, which are terms that are emphasized through audio or visual cues. These bonus words are used to enhance the relevance of the summary by highlighting words that carry additional semantic or emotional weight. Our approach generates both a concise textual summary and a timestamp-aligned video summary, ensuring that the audiovisual context is preserved. To direct our research, we propose the following research questions relevant to our proposed work.\\
% \newline
\textbf{(RQ1)} How can multimodal cues, such as audio prosody, visual expressions, and textual alignment, be effectively used to identify important events in video content?\\
\textbf{(RQ2)} How does the inclusion of \textit{bonus words}, emphasized through visual or auditory cues, impact the informativeness and relevance of generated summaries?\\
\textbf{(RQ3)} How does aligning transcript data with timestamped prosodic and visual cues improve the coherence and precision of both video and text-based summaries?

This work presents a novel approach to multimodal video summarization. Our contribution focuses on the integration of multimodal analysis, extending beyond static content extraction to capture audio, text and visual cues. The main contributions are as follows:
\begin{itemize}
    \item We propose a novel multimodal summarization framework that combines audio (prosodic features), visual (expressions and gestures), and textual (transcripts) cues to create timestamp-aligned textual and video summaries, addressing the limitations of traditional unimodal methods.
    \item We introduce a methodology for identifying \textit{bonus words}, terms that are emphasized through visual, audio, and textual cues. By prioritizing these words, we improve the emotional expressiveness and semantic accuracy of the generated summaries.
    \item We evaluate the generated summaries using both text-based and video-based metrics, providing a more comprehensive framework for assessing the quality and relevance of multimodal video summaries.
\end{itemize}

\section{Related Works}
\label{sec:2}

\noindent \textbf{Text Summarization.} Text summarization addresses the increasing demand for condensing large volumes of digital text into concise and meaningful summaries. Among the various summarization techniques, extractive summarization is particularly relevant to this work. This method selects key phrases directly from the original text, ensuring factual accuracy, but may struggle with maintaining smooth transitions and coherence between ideas \cite{wibawa2024survey}. A widely used approach within extractive summarization is the Edmundson heuristic method, which scores sentences based on features such as term frequency, sentence position, and cue phrases, allowing for the selection of the most important sentences \cite{edmundson1969new}. This technique is practical, systematic, and well-suited for the methodology employed in this study.

Although abstractive summarization, which generates new sentences for better fluency and readability, exists as an alternative, it is less relevant to the current focus and can sometimes introduce factual inaccuracies \cite{wibawa2024survey, 10826032, see2017get}. While recent advancements, particularly transformer models, have enhanced the effectiveness of both extractive and abstractive summarization, the Edmundson heuristic approach remains a reliable and effective foundation for extracting key information in this context.\\

\noindent \textbf{Video Summarization.} Traditional video summarization methods have primarily focused on extracting key frames or segments that represent the core content of the video. With the rise of machine learning and deep learning, modern approaches have increasingly relied on data-driven methods that emphasize feature extraction and semantic understanding. For instance, reinforcement learning has been used to improve the diversity and representativeness of video summaries \cite{zhou2018deep, liu2022video, li2021weakly, wang2024progressive}, while deep neural networks have shown promising results in capturing temporal relationships in video data \cite{zhang2016video, saini2023video}.

Multimodal video summarization has emerged as a promising solution to overcome the limitations of unimodal approaches by integrating visual, audio, and textual modalities. Early work by \citet{evangelopoulos2013multimodal} combined saliency cues from different modalities to identify key segments, but this approach lacked the ability to capture dynamic behavioural aspects, such as emotional tone or speaker gestures. More recent techniques have improved upon these methods by incorporating deep learning frameworks to better fuse information across modalities. \citet{park2022multimodal} proposed a transformer-based model integrating audio, visual, and textual features, while \citet{zhao2021audiovisual} introduced an audiovisual attention mechanism that aligns audio and visual cues. Recent studies have further advanced multimodal summarization through text integration. \citet{zhao2022hierarchical} introduced a Hierarchical Multimodal Transformer (HMT) that captures local and global dependencies across modalities. While HMT enhances summary coherence, it is primarily designed for professional video content and does not address the challenges of user-generated or conversational videos, where behavioural cues are more pronounced. 

Despite significant progress, existing approaches face limitations in capturing the full spectrum of human behaviour, such as emotional emphasis, facial expressions, or gesture-based communication, which are crucial for understanding speaker intent in conversational or emotive videos. While \citet{psallidas2021multimodal} focused on user-generated content, their method neglects behavioural cues essential for understanding speaker intent. Similarly, \citet{zhu2023topic} prioritized topic diversity but overlooked expressive elements like emotional emphasis or gesture-based communication. Also, external knowledge bases have been leveraged to enhance semantic understanding \cite{xie2024video, xie2022knowledge}, but their approach requires extensive external resources that may not be feasible for all video types.

In response to these limitations, there is a clear need for further research into multimodal video summarization that better capitalizes on the capabilities of automatic text summarization techniques. This work aims to fill these gaps by introducing a behaviour-aware multimodal summarization pipeline that focuses on identifying key elements in both the content and delivery of the video. By combining advanced techniques in audio-visual analysis and transcript alignment, we seek to generate timestamped video segments that not only represent the content of the video but also capture the expressive emphasis and emotional depth of the communication.

\section{Methodology}
\label{sec:3}

\begin{figure}[!h]
  \centering
  \includegraphics[width=\linewidth]{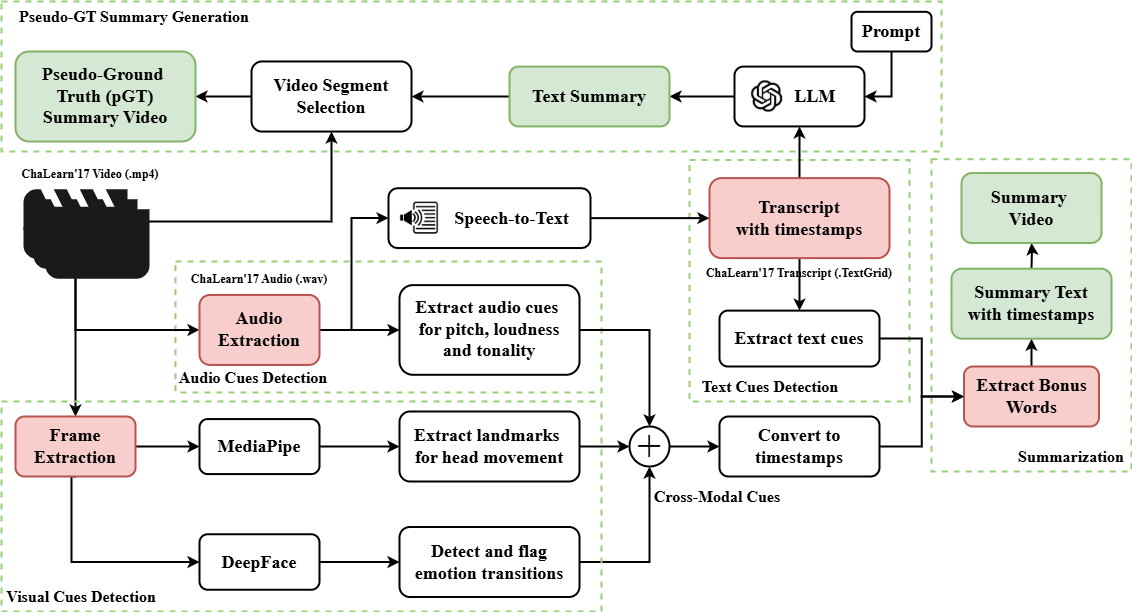}
  \caption{Multimodal Video Summarization Framework with Pseudo-Ground Truth Generation. The diagram illustrates a comprehensive pipeline for video summarization using the ChaLearn'17 dataset. Starting with input video, the process branches into parallel extraction paths: audio processing, frame extraction and speech-to-text conversion. The framework integrates audio, visual and textual cues extraction to generate a text summary with timestamps. This information is combined to select relevant video segments, producing both a standard summary video and LLM-prompted pseudo-ground truth (pGT) summary for evaluation purposes.}
  \label{fig:3}
\end{figure}

This section describes the proposed multimodal video summarization framework, which combines visual, audio and textual information to identify and extract key segments from the interview videos. The overall framework is illustrated in Figure \ref{fig:3}, encompassing data pre-processing, modality-specific cue extraction and summary generation.\\

% This study utilizes the ChaLearn First Impressions dataset \cite{ponce2016chalearn}, a publicly available collection of 10,000 short interview clips featuring diverse individuals speaking in English as part of their job interview task. The dataset captures rich multimodal behavioral cues, including facial expressions, gestures and speech patterns, making it ideal for human-centric video analysis. Originally designed for personality trait recognition and job interview score assessment, the dataset is repurposed in this work for multimodal video summarization. By integrating visual, audio and textual modalities, the proposed framework aims to generate concise summaries that preserve both semantic content and behavioral context. The dataset’s diversity and standardized format further support robust evaluation and generalizability of the summarization approach.\\

\noindent \textbf{Data Sources.} This study employs the ChaLearn First Impressions dataset \cite{ponce2016chalearn}, a publicly available collection of 10,000 high-quality video clips featuring 7,138 unique subjects speaking in English during short job interview tasks, averaging 15 seconds each (range: 8-20 seconds) at approximately 24 frames per second (fps). For our research, we selected a subset of 1,500 video clips (15\% of the total dataset). Originally designed for personality trait recognition and interview score assessment, it also captures diverse multimodal behavioural cues, such as facial expressions, gestures, and speech patterns, ideal for human-centric video analysis. This dataset is repurposed in this study for multimodal video summarization due to its’ standardized format and rich content, with transcripts averaging 38 words (about 152 characters) per clip, and it enables our framework to integrate visual, audio, and textual modalities to produce concise and meaningful summaries that preserve semantic and behavioural context. The diversity nature of the dataset support robust evaluation and generalizability of our approach.\\

\noindent \textbf{Audio Processing.} Audio processing and transcription are essential in our multimodal video summarization framework, as they allow for the accurate extraction and alignment of spoken content with other modalities. We start by separating the audio from video files using FFmpeg, ensuring high-quality, single-channel audio output with 16 kHz sampling rate. Next, we transcribe the audio using the Whisper large \cite{radford2022} automatic speech recognition (ASR) model with the language flag set to English, which performs well across different speech conditions, such as varying accents or background noise. To improve the transcripts’ clarity and structure, we apply natural language processing with SpaCy \cite{honnibal2020spacy}. This tool processes the raw transcripts by detecting sentence boundaries, adding proper punctuation, and refining the text into well-formed sentences, making it more suitable for summarization tasks. For precise timing, we use the Montreal Forced Aligner (MFA) \cite{mcauliffe2017montreal} to align each transcribed word with its exact time interval in the audio. This alignment links the spoken content to corresponding video frames and text, forming a strong basis for identifying key moments in the summarization process.\\

\begin{figure}
  \centering
  \includegraphics[width=\linewidth]{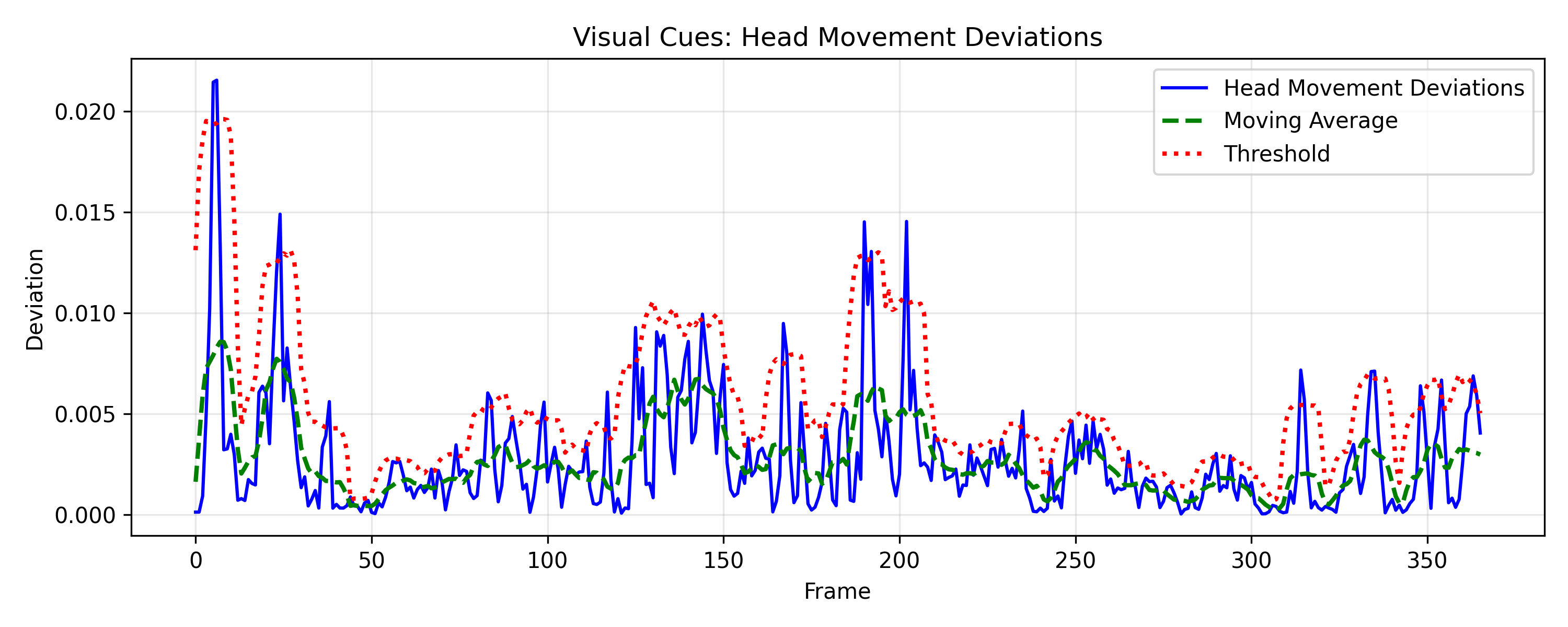}
  \caption{Example of head movement deviations over time, visualized as a function of video frames. The blue line represents frame-to-frame head movement deviations, the green dashed line shows the moving average of these deviations and the red dotted line indicates the dynamic threshold calculated as the moving average plus 1.5 times the standard deviation. Peaks above the threshold suggest significant head movement events, which may correspond to moments of increased visual salience.}
  \label{fig:1}
\end{figure}

\noindent \textbf{Visual Cues Detection.}Visual emphasis in our multimodal video summarization framework is derived from two key sources: head movement trajectories, as illustrated in Figure \ref{fig:1}, and facial emotion transitions. To ensure consistent analysis across videos, we extract frames at fixed intervals using OpenCV \cite{bradski2000opencv}. Each frame is processed with MediaPipe Pose \cite{lugaresi2019}, which detects 33 body landmarks. We select the nose landmark to represent head position. Head movement is calculated as the Euclidean displacement of the nose between consecutive frames. Any displacement exceeding a predefined threshold ($\lambda$) was flagged as a head movement cue, indicating physical emphasis or non-verbal communication.

% \begin{equation} 
% \text{Movement} = \sqrt{(x_t - x_{t-1})^2 + (y_t - y_{t-1})^2}
% \end{equation}

% This threshold was chosen empirically to capture visible shifts in head position, indicative of emphasis or gesture.
For facial emotions, we analyze each frame using the DeepFace library \cite{serengil2024lightface}, which employs multiple models for facial expression detection. The dominant emotion (e.g., happy, sad, angry) is identified per frame, and a visual cue is recorded when a transition occurs between distinct emotions. This approach highlights segments with emotional shifts, often tied to expressive storytelling or narrative peaks, enhancing the detection of key video moments.\\

\begin{figure}[h]
  \centering
  \includegraphics[width=\linewidth]{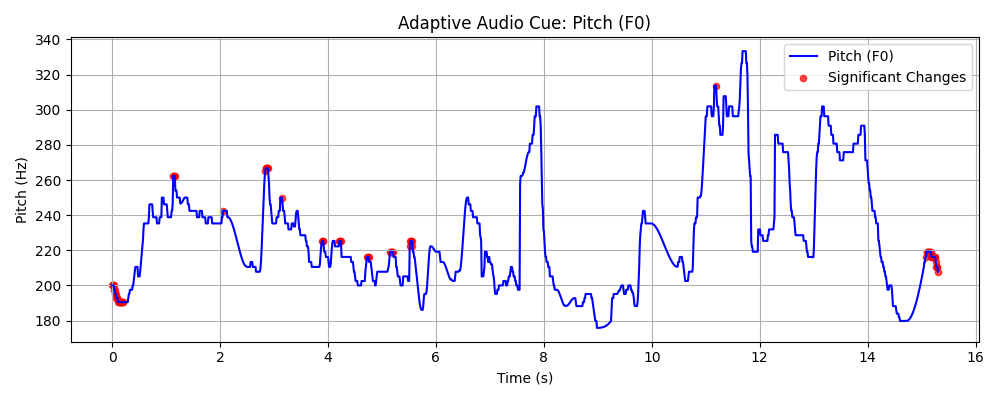}
  \caption{Example of pitch ($F_0$) variations over time. The blue line shows the interpolated pitch contour, and red dots mark significant changes based on a threshold, highlighting expressive moments relevant to summarization.}
  \label{fig:2}
\end{figure}

\noindent \textbf{Audio Cues Detection.} We process audio streams to detect prosodic emphasis through three key acoustic features: pitch, loudness, and tonality. Pitch ($F_0$) is extracted using the YAAPT algorithm \cite{kasi2002yet}, which effectively handles noise and unvoiced segments. We apply linear interpolation to missing values, creating a continuous time-aligned contour (Figure \ref{fig:2}). Loudness is quantified via short-time root mean square (RMS) energy, while voice quality is assessed using the \textit{HammarbergIndex}, the dB difference between the strongest harmonic peak in $0–2$ kHz and $2–5$ kHz ranges of the speech spectrum. This index characterizes spectral slope, with lower values indicating flatter spectra (suggesting vocal strain) and higher values reflecting greater low-frequency energy (associated with breathier voice). Both metrics are extracted using openSMILE \cite{eyben2010opensmile}. Each feature undergoes Z-score normalization:
\begin{equation}
z = \frac{x - \mu}{\sigma}
\end{equation}
where $x$ is the raw feature value, $\mu$ is the mean, and $\sigma$ is the standard deviation across the video. Significant audio cues are identified when normalized values exceed a predefined threshold, marking moments of prosodic emphasis that signal communicative importance.\\

\noindent \textbf{Textual Cues Detection.} Textual cues are extracted from transcribed audio to identify semantically and contextually significant elements in the dialogue, a key part of our multimodal video summarization framework. We use a multi-criteria approach to detect keywords based on statistical importance, emotional emphasis, and factual specificity. First, Term Frequency-Inverse Document Frequency (TF-IDF) highlights important terms, with common English stop-words removed to focus on meaningful words \cite{sparck1972statistical}. Sentiment analysis then identifies emotionally charged terms from high-intensity sentences, capturing affective content \cite{zhao2016sentiment}. Named Entity Recognition (NER) extracts key factual entities, such as people, organizations, and events, using SpaCy \cite{honnibal2020spacy}. Keywords from these methods are combined, filtered, and used to score sentences for the final summary. This approach ensures the summary includes the most significant, emotionally relevant, and fact-based elements of the transcript.\\

\textbf{Cross-Modal Cues Alignment.} Cross-modal alignment integrates visual, audio, and textual cues into a unified timeline for our video summarization framework. Visual cues, such as pose and emotion, are aligned to timestamps using video frame rates. Audio cues, timestamped during extraction, capture expressive features like pitch, tonality, and loudness. Textual cues are synchronized with audio using forced alignment techniques for word-level and sentence-level mapping. Sentence alignment is improved by matching word overlaps with time intervals, filtering out stopwords and duplicates. These aligned cues produce weighted textual \textit{bonus words} based on their links to key visual and audio events. A bonus word is selected when a textual term coincides temporally with salient cues in at least one additional modality. For instance, in the sentence: \textit{"I'm doing better than I was before, but I thought that comics are good because you get the words in the pictures,”} the words, such as \textit{good} (textual cues), \textit{pictures} and \textit{words} (visual cues), and \textit{comics} and \textit{pictures} (audio emphasis), are identified as \textit{bonus words} due to their multimodal prominence. This method builds on standard multimodal analysis to highlight semantically and behaviourally significant moments for summary generation.\\

% \noindent \textbf{Summarization \& Video Compilation.} The summarization phase ranks sentences from the transcript based on the density of multimodal cues, using an adaptive weighting mechanism informed by the presence of "bonus words" derived from aligned textual, audio and visual features to highlight key moments in the video. Sentences are evaluated using a weighting mechanism. These weights reflect the frequency of bonus words within each sentence, capturing their alignment with important multimodal events. To enhance summary quality, an adaptive selection process, incorporates both weight-based thresholding and diversity considerations, minimizing redundancy while preserving semantic richness. This process, detailed in Algorithm \ref{algo:1}, ensures that the selected segments accurately reflect semantically and behaviorally salient moments within the video. The weighting process assigns scores to sentences proportional to the occurrence of bonus words, a method similar to Edmundson summarization \cite{edmundson1969new}. An adaptive threshold, calculated as the mean weight plus a factor of the standard deviation, filters out sentences with sufficient salience. To ensure diversity, a penalty is applied based on textual similarity between selected sentences, using cosine similarity derived from term frequency-inverse document frequency (TF-IDF) quantification. Sentences with excessive overlap are therefore excluded unless their similarity penalty score falls below a predefined limit, balancing informativeness and conciseness. 

\textbf{Summarization \& Video Compilation.} In the summarization phase, we rank transcript sentences based on multimodal cue density, using a weighting method that highlights \textit{bonus words} from aligned textual, audio, and visual features, following Edmundson’s summarization approach \cite{edmundson1969new}. Sentence weights are set by the frequency of \textit{bonus words}, marking important video moments. An adaptive selection process, detailed in Algorithm \ref{algo:1}, filters sentences with a weight-based threshold and promotes diversity. The threshold uses the mean weight and standard deviation. To avoid repetition, a penalty based on cosine similarity from TF-IDF scores is applied \cite{sparck1972statistical}. Sentences with high overlap are excluded unless their similarity penalty is low. This ensures a concise yet informative summary.

\begin{algorithm}
\caption{Multimodal Video Summarization}
\label{algo:1}
\begin{algorithmic}[1]
\Require Video segments $V = {V_1, \ldots, V_N}$, Transcript $T$, threshold $\lambda$, diversity factor $\delta$
\Ensure Summary $S$

\State $W_{combined} \gets \emptyset$
\For{each $V_i$ in $V$}
    \State Extract textual, audio, and visual cues: $T_i, A_i, \mathcal{V}i$
    \State Extract significant words from each modality
    \State $W_{combined} \gets W_{combined} \cup$ words from all modalities
\EndFor

\State $S_{all} \gets$ Split($T$) \Comment{Split transcript into sentences}
\State $W_s \gets \emptyset$ \Comment{Initialize sentence weights}
\For{each $s$ in $S_{all}$}
    \State $w_s \gets 0$
    \State Words $\gets$ Tokenize(Lowercase($s$))
    \For{each $w$ in $W_{combined}$}
        \State $w_s \gets w_s + \text{Count}(w, \text{Words})$
    \EndFor
    \State $W_s[s] \gets w_s$
\EndFor

\State $\theta \gets$ Mean(Values($W_s$)) $+ \lambda \cdot$ StdDev(Values($W_s$)) \Comment{Adaptive threshold}
\State $C \gets \{s \in S_{all} \mid W_s[s] \geq \theta\}$ \Comment{Candidate sentences}
\State $S_{final} \gets \emptyset$ \Comment{Initialize selected sentences}

\For{each $s$ in $C$}
    \If{$S_{final}$ is empty}
        \State Add $s$ to $S_{final}$
    \Else
        \State $sim \gets$ MaxCosineSimilarity($s$, $S_{final}$, TF-IDF)
        \If{$sim \cdot \delta < 0.5$} \Comment{Diversity check}
            \State Add $s$ to $S_{final}$
        \EndIf
    \EndIf
\EndFor

\State \Return $S \gets$ Join($S_{final}$, '\ ') \Comment{Final summary}
\end{algorithmic}
\end{algorithm}

Selected segments are converted to frame ranges using the video’s frame rate and extracted with FFmpeg. Subtitles are added by aligning text to time intervals, and segments are combined into a single summary video with embedded subtitles. The final output is a temporally aligned, content-rich video summary that integrates multimodal cues, offering an interpretable and compressed representation of the original content.

\section{Experiment}
\label{sec:4}

% Our proposed multimodal video summarization framework processes videos at a sampling rate of 1 frame per second. Audio is extracted using FFmpeg, transcribed with Whisper \cite{radford2022} and aligned with Montreal Forced Aligner (MFA) \cite{mcauliffe2017montreal} for word-level timestamps. Visual cues are detected using MediaPipe and DeepFace \cite{lugaresi2019, serengil2024lightface}, while audio features are extracted with openSMILE \cite{eyben2010opensmile} and YAAPT. Textual cues, including \textit{bonus words}, are identified using TF-IDF, sentiment analysis and NER, as outlined in Section \ref{sec:3}. The summarization process, formalized in Algorithm \ref{algo:1}, weights sentences based on bonus word frequency, applies adaptive thresholding ($λ=0.3$) and ensures diversity ($δ=0.2$) using TF-IDF and cosine similarity. Timestamped segments are mapped, extracted with FFmpeg and concatenated into a summary video.\\

\noindent \textbf{Implementation Details.} Our multimodal video summarization framework samples videos at 1 frame per second. Audio is extracted with FFmpeg, transcribed using Whisper \cite{radford2022}, and refined with SpaCy for sentence structure \cite{honnibal2020spacy}. Word-level timestamps are obtained using the Montreal Forced Aligner \cite{mcauliffe2017montreal}. Visual cues are detected with MediaPipe for pose and DeepFace for emotions \cite{lugaresi2019, serengil2024lightface}. Audio features, such as pitch, are extracted using openSMILE and YAAPT \cite{eyben2010opensmile, kasi2002yet}. Textual cues, including bonus words, are identified with TF-IDF, sentiment analysis, and NER, as described in Section \ref{sec:3}. Sentences are weighted by bonus word frequency, filtered with an adaptive threshold, and diversified ($\delta=0.2$) using TF-IDF and cosine similarity. The adaptive threshold is as follows,

\begin{equation}
\theta = \mu + \lambda \cdot \sigma
\end{equation}

where $\mu$ and $\sigma$ are the mean and standard deviation of the sentence weights across all terms, and $\lambda$ is a threshold factor of 0.3. Timestamped segments are extracted and combined into a summary video with subtitles.

\subsection{Pseudo-Ground Truth Summary Generation} 
\label{sec:4.1}
Manual annotation for video summarization is expensive, subjective, and often infeasible due to the unavailability of human-annotated data for large-scale datasets. To overcome this, we adopt a scalable extractive summarization approach, inspired by \citet{argaw2024scaling} and TL;DW? \cite{narasimhan2022tl}, leveraging the text comprehension capabilities of large language models (LLMs) to generate pseudo-ground truth (pGT) summaries, as illustrated in Figure \ref{fig:3}. 

We use the ChaLearn First Impressions dataset, previously mentioned in Section \ref{sec:3}, where each video features a single speaker in a controlled setting, enabling clean audio and strong speech-to-video alignment. Audio is then extracted and transcribed using Whisper. Sentence boundaries and punctuation are refined for improved readability. Word-level timestamps are aligned using the Montreal Forced Aligner (MFA) \cite{mcauliffe2017montreal}, producing a timestamped transcript where each sentence is tagged with its start and end time (e.g., \texttt{[7.21s, 8.51s] This is my thing.}). To generate summaries, we use GPT-4.5 model with a prompt-tuned extractive strategy. The model selects only the most critical and informative sentences without modifying their wording. Timestamps are preserved to allow direct mapping to video segments. The final summary video is created by extracting and concatenating these segments chronologically. The resulting pGT summaries provide a consistent, text-driven reference summary for evaluating our proposed multimodal summarization framework. The LLM prompt is shown below,

\begin{tcolorbox}[colback=white, colframe=gray!30]
\textit{You are given a transcript of a spoken video with sentence-level timestamps. Your task is to create an extractive summary by selecting only the most informative and important sentences from the transcript.\\
Instructions:\\
1. Do not paraphrase or modify the wording. Retain the exact original sentence.\\
2. Include only the most critical and informative sentences.\\
3. Preserve the timestamps for each selected sentence.\\
4. Return the output as a list of [start\_time, end\_time, sentence].}\\ 
\textbf{Transcript:} [Transcript]
% \textbf{Summary:}
\end{tcolorbox}

We acknowledge the limitations of pGT evaluation, particularly potential biases in the LLM's sentence selection and generalizability concerns. To address these limitations while working within our study constraints, we implemented a two-part validation strategy. First, we conducted a rigorous qualitative analysis on a random subset of 50 generated summaries, manually verifying that selected sentences capture both key content and essential behavioural cues. Our analysis revealed that around 93\% of examined summaries effectively represented primary themes while preserving important prosodic and visual emphasis points.
Second, we performed consistency verification by generating multiple pGT versions for a sample of videos under different prompt conditions. The high agreement between versions (average Jaccard similarity of 0.78) suggests stability in our pGT generation approach.

While our validation approach cannot fully substitute human-annotated ground truth, it provides reasonable assurance that our pGT summaries capture meaningful content and behavioural elements. The consistency of our results across multiple evaluation metrics further supports the reliability of our findings, though we acknowledge the need for future work to explore additional validation methods for enhanced generalizability across diverse video contexts.

\subsection{Experimental Results}
\label{sec:4.2}

We evaluate our proposed multimodal summarization framework by comparing its generated text and video summaries against pseudo-ground truth (pGT) LLM-generated summaries described in Section \ref{sec:4.1}. These timestamped pGT summaries serve as scalable references derived directly from the dataset, enabling comprehensive evaluation of our framework's ability to enhance summarization through multimodal integration.\\

\noindent \textbf{Text-Based Evaluation.} We compare summaries from our proposed multimodal framework against the baseline Edmundson summarizer using pGT references and multiple complementary metrics. ROUGE-1, ROUGE-2, and ROUGE-L assess unigram overlap, bigram overlap, and longest common subsequence respectively, capturing lexical and structural alignment. BLEU quantifies n-gram precision with a brevity penalty to measure reference fidelity, while BERTScore employs contextual embeddings to evaluate semantic similarity beyond surface forms. All metrics are computed per summary and averaged across the dataset. Table \ref{tab:1} demonstrates the substantial performance differences between approaches. The Edmundson summarizer produces concise summaries (length ratio: 0.3125) that capture approximately 48\% of pGT content, with modest BLEU (0.2259) and relatively high BERTScore (0.9152) values indicating the limitations of its purely extractive approach. In contrast, our multimodal framework consistently outperforms across all metrics. With a length ratio of 0.5734, it achieves substantially higher ROUGE scores, a significant improvement in BLEU score (0.6411) and enhanced semantic fidelity  with BERTScore of 0.9536.\\

\begin{table}[!h]
\centering
\caption{Comparative evaluation of text-based summaries against pseudo-ground truth (pGT) summaries. The proposed multimodal approach demonstrates substantial improvements across all metrics.}
\label{tab:1}
\begin{tabular}{lcccccc}
\toprule
\textbf{Method} & \textbf{Length Ratio} & \textbf{ROUGE-1} & \textbf{ROUGE-2} & \textbf{ROUGE-L} & \textbf{BLEU} & \textbf{BERTScore} \\ 
\midrule
Edmundson & 0.3125 & 0.4769 & 0.4496 & 0.4738 & 0.2259 & 0.9152 \\
Proposed (Multimodal) & \textbf{0.5734} & \textbf{0.7929} & \textbf{0.7763} & \textbf{0.7889} & \textbf{0.6411} & \textbf{0.9536} \\ 
\bottomrule
\end{tabular}
\end{table}

\noindent \textbf{Video-Based Evaluation.} We assess summary alignment and temporal consistency using three metrics that capture distinct quality aspects. F1-score measures segment selection accuracy through temporal intersection-over-union (IoU) between framework-generated and pGT segments, with matches defined by IoU > 0.5. Frame-level precision and recall (1 fps) are averaged across the dataset. Kendall's $\tau$ evaluates temporal sequence ordinal agreement by analyzing concordant and discordant pairs, while Spearman's $\rho$ quantifies monotonic correlation between ranked segment sequences, together assessing narrative coherence preservation. Table \ref{tab:2} shows our multimodal approach significantly outperforming the Edmundson baseline. The proposed method achieves a 22.5\% higher F1-score (0.6995 vs. 0.5709) and substantially improved temporal consistency measures (Kendall's $\tau$: 0.3148 vs. 0.2295; Spearman's $\rho$: 0.3690 vs. 0.2681). These improvements stem from the multimodal framework's enhanced ability to capture semantically relevant content across modalities. However, both methods' moderate rank correlations ($< 0.4$) suggest remaining challenges in replicating psuedo-ground truth narrative structure, likely due to timestamp mapping constraints and fixed sentence selection parameters.

\begin{table}[!h]
\centering
\caption{Comparative evaluation of summary videos against pseudo-ground truth (pGT) summary videos. The proposed multimodal approach demonstrates consistent improvements across all metrics, outperforming Edmundson in segment selection accuracy (F1-Score) and temporal coherence measures (Kendall's $\tau$ and Spearman's $\rho$).}
\label{tab:2}
\begin{tabular}{lccc}
\toprule
\textbf{Method} & \textbf{F1-Score} & \textbf{Kendall's $\tau$} & \textbf{Spearman's $\rho$} \\
\midrule
Edmundson & 0.5709 & 0.2295 & 0.2681 \\
Proposed (Multimodal) & \textbf{0.6995} & \textbf{0.3148} & \textbf{0.3690} \\
\bottomrule
\end{tabular}
\end{table}

This comprehensive evaluation results demonstrate the clear advantages of our multimodal approach. By integrating audio, textual and visual cues, our framework achieves an optimal balance between summary conciseness and content comprehensiveness. The significant improvements across all evaluation metrics position our proposed multimodal framework as a substantial advancement over traditional extractive techniques. These results highlight the framework's ability to capture more nuanced and complete representations of source content, validating the efficacy of multimodal integration for video summarization tasks.

\subsection{Ablation Study}
\label{sec:4.3}
In Table \ref{tab:3}, we perform ablation experiments to evaluate key components of our multimodal video summarization framework. Each variant is evaluated against pseudo-ground truth (pGT) LLM-generated summaries on the ChaLearn First Impression dataset.

\begin{table}[!h]
\scriptsize
\centering
\caption{Ablation studies on the ChaLearn First Impression dataset.}
\label{tab:3}
\begin{tabular}{lcccccccc}
\toprule
\textbf{Method} & \textbf{ROUGE-1} & \textbf{ROUGE-2} & \textbf{ROUGE-L} & \textbf{BLEU} & \textbf{BERTScore} & \textbf{F1-Score} & \textbf{Kendall's $\tau$} & \textbf{Spearman's $\rho$} \\
\midrule
Text & 0.5214 & 0.4876 & 0.5132 & 0.5823 & 0.9178 & 0.5496 & 0.2765 & 0.3123 \\
Audio & 0.3921 & 0.3456 & 0.3874 & 0.4312 & 0.8634 & 0.5843 & 0.1956 & 0.2345 \\
Visual & 0.6743 & 0.6321 & 0.6689 & 0.7234 & 0.9412 & 0.6789 & 0.2987 & 0.3456 \\
w/o Audio & 0.7123 & 0.6894 & 0.7056 & \textbf{0.7456} & 0.9489 & 0.6876 & 0.3054 & 0.3578 \\
w/o Visual & 0.4567 & 0.4123 & 0.4498 & 0.4978 & 0.8923 & 0.6234 & 0.2345 & 0.2789 \\
w/o Text & 0.7456 & 0.7234 & 0.7389 & 0.6895 & 0.9501 & 0.6923 & 0.3098 & 0.3621 \\
\midrule
\textbf{Multimodal} & \textbf{0.7929} & \textbf{0.7763} & \textbf{0.7889} & 0.6411 & \textbf{0.9536} & \textbf{0.6995} & \textbf{0.3148} & \textbf{0.3690} \\
\bottomrule
\end{tabular}
\end{table}

We examine the role of textual information by removing text and relying solely on audio-visual inputs. As shown in Table \ref{tab:3}, this variant without text cues yields reasonable results. However, integrating textual context substantially enhances summarization quality, emphasizing the benefit of multimodal learning over an audio-visual only framework. To assess the contribution of audio features, we omit audio cues, using only text and visual cues. While this approach maintains fair performance, incorporating audio information clearly provides complementary context, underscoring audio’s supporting role in capturing summarization-relevant nuances. Removing visual modality significantly reduces model effectiveness, highlighting visual features as critical for accurate video summarization. This outcome confirms the essential role of visual data in understanding semantic and structural aspects of videos. Next, evaluating each modality independently shows visual information providing the strongest single-modality performance, followed by text and audio. All single-modality variants clearly underperform compared to multimodal approaches, reinforcing the importance of multimodal integration. Our integrated multimodal approach, which combines text, audio, and visual features through cross-modal learning, consistently achieves the best results across all metrics except BLEU, likely due to BLEU favouring exact matches over semantic depth. Our full multimodal approach excels, with BERTScore at 0.9536, validating that integrating all modalities enhances summarization quality. These findings validate our hypothesis that leveraging multimodal context significantly enhances video summarization effectiveness.

\section{Conclusion} 
\label{sec:5}

This study presents a scalable and a novel multimodal video summarization framework that integrates visual, audio, and textual cues to generate concise and meaningful summaries from the ChaLearn First Impressions dataset. Unlike traditional unimodal approaches, our method leverages behavioural and prosodic signals, such as head movements, facial emotion transitions, pitch, loudness, and word emphasis, to capture key moments and the full communicative intent of the speaker in interview videos. We introduce a mechanism to detect and weight \textit{bonus words}, which are semantically or emotionally salient based on cross-modal alignment, thereby improving both the informativeness and relevance of the generated summaries. To overcome the challenge of limited human-annotated data, we adopt a pseudo-ground truth (pGT) generation strategy using LLM-based extractive summarization, allowing us to evaluate our framework against consistent reference summaries.

Our framework offers practical applications across multiple domains: transforming educational lectures into concise summaries that preserve instructor emphasis; extracting insights from professional meetings while maintaining emotional context; helping content creators identify engaging moments for previews; and enhancing accessibility for users with time constraints or cognitive challenges. These applications demonstrate how our behaviour-aware approach to video summarization addresses real-world needs for efficient video content consumption while preserving communicative nuance.

Experimental results demonstrate that our approach significantly outperforms both unimodal baselines and traditional extractive summarizers such as the Edmundson method, across a wide range of text and video-based metrics. Ablation studies confirm that while visual features provide the strongest individual contribution, integrating all modalities yields optimal performance. These findings highlight the importance of multimodal fusion for accurate and expressive video summarization. Future work could explore integrating large vision-language models for end-to-end training, expanding the framework to longer-form videos, and incorporating personalized summarization techniques to further improve narrative coherence and adaptability across diverse datasets. This study underscores the value of behaviour-aware, multimodal understanding in bridging the gap between human communication and machine-generated summaries.

\section*{Declaration on Generative AI}
 %  {\em Either:}\newline
 %  The author(s) have not employed any Generative AI tools.
 % %  \newline
  
 % \noindent{\em Or (by using the activity taxonomy in ceur-ws.org/genai-tax.html):\newline}
 During the preparation of this work, the author(s) used Grammarly in order to: perform grammar, style, and spelling check, and to improve words and phrases in some sections. After using the tool, the author(s) reviewed and edited the content as needed and take(s) full responsibility for the publication’s content. 

\section*{Acknowledgment}
This research is partially funded by the University of Oulu and the Finnish Research Council Flagship 7 -Hybrid Intelligence-, which is gratefully acknowledged. The first author is funded from the Finnish Research Council Digital Water Flagship project.

%%
%% Define the bibliography file to be used
\bibliography{bibliography}

\begin{thebibliography}{36}
\expandafter\ifx\csname natexlab\endcsname\relax\def\natexlab#1{#1}\fi
\providecommand{\url}[1]{\texttt{#1}}
\providecommand{\href}[2]{#2}
\providecommand{\path}[1]{#1}
\providecommand{\DOIprefix}{doi:}
\providecommand{\ArXivprefix}{arXiv:}
\providecommand{\URLprefix}{URL: }
\providecommand{\Pubmedprefix}{pmid:}
\providecommand{\doi}[1]{\href{http://dx.doi.org/#1}{\path{#1}}}
\providecommand{\Pubmed}[1]{\href{pmid:#1}{\path{#1}}}
\providecommand{\bibinfo}[2]{#2}
\ifx\xfnm\relax \def\xfnm[#1]{\unskip,\space#1}\fi
%Type = Article
\bibitem[{Apostolidis et~al.(2021)Apostolidis, Adamantidou, Metsai, Mezaris, and Patras}]{apostolidis2021video}
\bibinfo{author}{E.~Apostolidis}, \bibinfo{author}{E.~Adamantidou}, \bibinfo{author}{A.~I. Metsai}, \bibinfo{author}{V.~Mezaris}, \bibinfo{author}{I.~Patras},
\newblock \bibinfo{title}{Video summarization using deep neural networks: A survey},
\newblock \bibinfo{journal}{Proceedings of the IEEE} \bibinfo{volume}{109} (\bibinfo{year}{2021}) \bibinfo{pages}{1838--1863}.
%Type = Inproceedings
\bibitem[{Chu et~al.(2015)Chu, Song, and Jaimes}]{chu2015video}
\bibinfo{author}{W.-S. Chu}, \bibinfo{author}{Y.~Song}, \bibinfo{author}{A.~Jaimes},
\newblock \bibinfo{title}{Video co-summarization: Video summarization by visual co-occurrence},
\newblock in: \bibinfo{booktitle}{Proceedings of the IEEE conference on computer vision and pattern recognition}, \bibinfo{year}{2015}, pp. \bibinfo{pages}{3584--3592}.
%Type = Article
\bibitem[{Tiwari and Bhatnagar(2021)}]{tiwari2021survey}
\bibinfo{author}{V.~Tiwari}, \bibinfo{author}{C.~Bhatnagar},
\newblock \bibinfo{title}{A survey of recent work on video summarization: approaches and techniques},
\newblock \bibinfo{journal}{Multimedia Tools and Applications} \bibinfo{volume}{80} (\bibinfo{year}{2021}) \bibinfo{pages}{27187--27221}.
%Type = Inproceedings
\bibitem[{Otani et~al.(2017)Otani, Nakashima, Rahtu, Heikkil{\"a}, and Yokoya}]{otani2017video}
\bibinfo{author}{M.~Otani}, \bibinfo{author}{Y.~Nakashima}, \bibinfo{author}{E.~Rahtu}, \bibinfo{author}{J.~Heikkil{\"a}}, \bibinfo{author}{N.~Yokoya},
\newblock \bibinfo{title}{Video summarization using deep semantic features},
\newblock in: \bibinfo{booktitle}{Computer Vision--ACCV 2016: 13th Asian Conference on Computer Vision, Taipei, Taiwan, November 20-24, 2016, Revised Selected Papers, Part V 13}, \bibinfo{organization}{Springer}, \bibinfo{year}{2017}, pp. \bibinfo{pages}{361--377}.
%Type = Inproceedings
\bibitem[{Rochan et~al.(2018)Rochan, Ye, and Wang}]{rochan2018video}
\bibinfo{author}{M.~Rochan}, \bibinfo{author}{L.~Ye}, \bibinfo{author}{Y.~Wang},
\newblock \bibinfo{title}{Video summarization using fully convolutional sequence networks},
\newblock in: \bibinfo{booktitle}{Proceedings of the European conference on computer vision (ECCV)}, \bibinfo{year}{2018}, pp. \bibinfo{pages}{347--363}.
%Type = Article
\bibitem[{Wibawa et~al.(2024)Wibawa, Kurniawan et~al.}]{wibawa2024survey}
\bibinfo{author}{A.~P. Wibawa}, \bibinfo{author}{F.~Kurniawan}, et~al.,
\newblock \bibinfo{title}{A survey of text summarization: Techniques, evaluation and challenges},
\newblock \bibinfo{journal}{Natural Language Processing Journal} \bibinfo{volume}{7} (\bibinfo{year}{2024}) \bibinfo{pages}{100070}.
%Type = Article
\bibitem[{Edmundson(1969)}]{edmundson1969new}
\bibinfo{author}{H.~P. Edmundson},
\newblock \bibinfo{title}{New methods in automatic extracting},
\newblock \bibinfo{journal}{Journal of the ACM (JACM)} \bibinfo{volume}{16} (\bibinfo{year}{1969}) \bibinfo{pages}{264--285}.
%Type = Inproceedings
\bibitem[{Islam et~al.(2024)Islam, Muhammad, and Oussalah}]{10826032}
\bibinfo{author}{M.~M. Islam}, \bibinfo{author}{U.~Muhammad}, \bibinfo{author}{M.~Oussalah},
\newblock \bibinfo{title}{Evaluating text summarization techniques and factual consistency with language models},
\newblock in: \bibinfo{booktitle}{2024 IEEE International Conference on Big Data (BigData)}, \bibinfo{year}{2024}, pp. \bibinfo{pages}{116--122}. \DOIprefix\doi{10.1109/BigData62323.2024.10826032}.
%Type = Article
\bibitem[{See et~al.(2017)See, Liu, and Manning}]{see2017get}
\bibinfo{author}{A.~See}, \bibinfo{author}{P.~J. Liu}, \bibinfo{author}{C.~D. Manning},
\newblock \bibinfo{title}{Get to the point: Summarization with pointer-generator networks},
\newblock \bibinfo{journal}{arXiv preprint arXiv:1704.04368}  (\bibinfo{year}{2017}).
%Type = Inproceedings
\bibitem[{Zhou et~al.(2018)Zhou, Qiao, and Xiang}]{zhou2018deep}
\bibinfo{author}{K.~Zhou}, \bibinfo{author}{Y.~Qiao}, \bibinfo{author}{T.~Xiang},
\newblock \bibinfo{title}{Deep reinforcement learning for unsupervised video summarization with diversity-representativeness reward},
\newblock in: \bibinfo{booktitle}{Proceedings of the AAAI conference on artificial intelligence}, volume~\bibinfo{volume}{32}, \bibinfo{year}{2018}.
%Type = Article
\bibitem[{Liu et~al.(2022)Liu, Meng, Huang, Vlontzos, Rueckert, and Kainz}]{liu2022video}
\bibinfo{author}{T.~Liu}, \bibinfo{author}{Q.~Meng}, \bibinfo{author}{J.-J. Huang}, \bibinfo{author}{A.~Vlontzos}, \bibinfo{author}{D.~Rueckert}, \bibinfo{author}{B.~Kainz},
\newblock \bibinfo{title}{Video summarization through reinforcement learning with a 3d spatio-temporal u-net},
\newblock \bibinfo{journal}{IEEE transactions on image processing} \bibinfo{volume}{31} (\bibinfo{year}{2022}) \bibinfo{pages}{1573--1586}.
%Type = Inproceedings
\bibitem[{Li and Yang(2021)}]{li2021weakly}
\bibinfo{author}{Z.~Li}, \bibinfo{author}{L.~Yang},
\newblock \bibinfo{title}{Weakly supervised deep reinforcement learning for video summarization with semantically meaningful reward},
\newblock in: \bibinfo{booktitle}{Proceedings of the IEEE/CVF winter conference on applications of computer vision}, \bibinfo{year}{2021}, pp. \bibinfo{pages}{3239--3247}.
%Type = Article
\bibitem[{Wang et~al.(2024)Wang, Wu, and Yan}]{wang2024progressive}
\bibinfo{author}{G.~Wang}, \bibinfo{author}{X.~Wu}, \bibinfo{author}{J.~Yan},
\newblock \bibinfo{title}{Progressive reinforcement learning for video summarization},
\newblock \bibinfo{journal}{Information Sciences} \bibinfo{volume}{655} (\bibinfo{year}{2024}) \bibinfo{pages}{119888}.
%Type = Inproceedings
\bibitem[{Zhang et~al.(2016)Zhang, Chao, Sha, and Grauman}]{zhang2016video}
\bibinfo{author}{K.~Zhang}, \bibinfo{author}{W.-L. Chao}, \bibinfo{author}{F.~Sha}, \bibinfo{author}{K.~Grauman},
\newblock \bibinfo{title}{Video summarization with long short-term memory},
\newblock in: \bibinfo{booktitle}{Computer Vision--ECCV 2016: 14th European Conference, Amsterdam, The Netherlands, October 11--14, 2016, Proceedings, Part VII 14}, \bibinfo{organization}{Springer}, \bibinfo{year}{2016}, pp. \bibinfo{pages}{766--782}.
%Type = Article
\bibitem[{Saini et~al.(2023)Saini, Kumar, Kashid, Saini, and Negi}]{saini2023video}
\bibinfo{author}{P.~Saini}, \bibinfo{author}{K.~Kumar}, \bibinfo{author}{S.~Kashid}, \bibinfo{author}{A.~Saini}, \bibinfo{author}{A.~Negi},
\newblock \bibinfo{title}{Video summarization using deep learning techniques: a detailed analysis and investigation},
\newblock \bibinfo{journal}{Artificial Intelligence Review} \bibinfo{volume}{56} (\bibinfo{year}{2023}) \bibinfo{pages}{12347--12385}.
%Type = Article
\bibitem[{Evangelopoulos et~al.(2013)Evangelopoulos, Zlatintsi, Potamianos, Maragos, Rapantzikos, Skoumas, and Avrithis}]{evangelopoulos2013multimodal}
\bibinfo{author}{G.~Evangelopoulos}, \bibinfo{author}{A.~Zlatintsi}, \bibinfo{author}{A.~Potamianos}, \bibinfo{author}{P.~Maragos}, \bibinfo{author}{K.~Rapantzikos}, \bibinfo{author}{G.~Skoumas}, \bibinfo{author}{Y.~Avrithis},
\newblock \bibinfo{title}{Multimodal saliency and fusion for movie summarization based on aural, visual, and textual attention},
\newblock \bibinfo{journal}{IEEE Transactions on Multimedia} \bibinfo{volume}{15} (\bibinfo{year}{2013}) \bibinfo{pages}{1553--1568}.
%Type = Article
\bibitem[{Park et~al.(2022)Park, Kwoun, Lee, and Lim}]{park2022multimodal}
\bibinfo{author}{J.~Park}, \bibinfo{author}{K.~Kwoun}, \bibinfo{author}{C.~Lee}, \bibinfo{author}{H.~Lim},
\newblock \bibinfo{title}{Multimodal frame-scoring transformer for video summarization},
\newblock \bibinfo{journal}{arXiv preprint arXiv:2207.01814}  (\bibinfo{year}{2022}).
%Type = Article
\bibitem[{Zhao et~al.(2021)Zhao, Gong, and Li}]{zhao2021audiovisual}
\bibinfo{author}{B.~Zhao}, \bibinfo{author}{M.~Gong}, \bibinfo{author}{X.~Li},
\newblock \bibinfo{title}{Audiovisual video summarization},
\newblock \bibinfo{journal}{IEEE Transactions on Neural Networks and Learning Systems} \bibinfo{volume}{34} (\bibinfo{year}{2021}) \bibinfo{pages}{5181--5188}.
%Type = Article
\bibitem[{Zhao et~al.(2022)Zhao, Gong, and Li}]{zhao2022hierarchical}
\bibinfo{author}{B.~Zhao}, \bibinfo{author}{M.~Gong}, \bibinfo{author}{X.~Li},
\newblock \bibinfo{title}{Hierarchical multimodal transformer to summarize videos},
\newblock \bibinfo{journal}{Neurocomputing} \bibinfo{volume}{468} (\bibinfo{year}{2022}) \bibinfo{pages}{360--369}.
%Type = Article
\bibitem[{Psallidas et~al.(2021)Psallidas, Koromilas, Giannakopoulos, and Spyrou}]{psallidas2021multimodal}
\bibinfo{author}{T.~Psallidas}, \bibinfo{author}{P.~Koromilas}, \bibinfo{author}{T.~Giannakopoulos}, \bibinfo{author}{E.~Spyrou},
\newblock \bibinfo{title}{Multimodal summarization of user-generated videos},
\newblock \bibinfo{journal}{Applied Sciences} \bibinfo{volume}{11} (\bibinfo{year}{2021}) \bibinfo{pages}{5260}.
%Type = Article
\bibitem[{Zhu et~al.(2023)Zhu, Zhao, Hua, and Wu}]{zhu2023topic}
\bibinfo{author}{Y.~Zhu}, \bibinfo{author}{W.~Zhao}, \bibinfo{author}{R.~Hua}, \bibinfo{author}{X.~Wu},
\newblock \bibinfo{title}{Topic-aware video summarization using multimodal transformer},
\newblock \bibinfo{journal}{Pattern Recognition} \bibinfo{volume}{140} (\bibinfo{year}{2023}) \bibinfo{pages}{109578}.
%Type = Article
\bibitem[{Xie et~al.(2024)Xie, Chen, Zhao, and Lu}]{xie2024video}
\bibinfo{author}{J.~Xie}, \bibinfo{author}{X.~Chen}, \bibinfo{author}{S.~Zhao}, \bibinfo{author}{S.-P. Lu},
\newblock \bibinfo{title}{Video summarization via knowledge-aware multimodal deep networks},
\newblock \bibinfo{journal}{Knowledge-Based Systems} \bibinfo{volume}{293} (\bibinfo{year}{2024}) \bibinfo{pages}{111670}.
%Type = Inproceedings
\bibitem[{Xie et~al.(2022)Xie, Chen, Lu, and Yang}]{xie2022knowledge}
\bibinfo{author}{J.~Xie}, \bibinfo{author}{X.~Chen}, \bibinfo{author}{S.-P. Lu}, \bibinfo{author}{Y.~Yang},
\newblock \bibinfo{title}{A knowledge augmented and multimodal-based framework for video summarization},
\newblock in: \bibinfo{booktitle}{Proceedings of the 30th ACM International Conference on Multimedia}, \bibinfo{year}{2022}, pp. \bibinfo{pages}{740--749}.
%Type = Inproceedings
\bibitem[{Ponce-L{\'o}pez et~al.(2016)Ponce-L{\'o}pez, Chen, Oliu, Corneanu, Clap{\'e}s, Guyon, Bar{\'o}, Escalante, and Escalera}]{ponce2016chalearn}
\bibinfo{author}{V.~Ponce-L{\'o}pez}, \bibinfo{author}{B.~Chen}, \bibinfo{author}{M.~Oliu}, \bibinfo{author}{C.~Corneanu}, \bibinfo{author}{A.~Clap{\'e}s}, \bibinfo{author}{I.~Guyon}, \bibinfo{author}{X.~Bar{\'o}}, \bibinfo{author}{H.~J. Escalante}, \bibinfo{author}{S.~Escalera},
\newblock \bibinfo{title}{Chalearn lap 2016: First round challenge on first impressions-dataset and results},
\newblock in: \bibinfo{booktitle}{Computer Vision--ECCV 2016 Workshops: Amsterdam, The Netherlands, October 8-10 and 15-16, 2016, Proceedings, Part III 14}, \bibinfo{organization}{Springer}, \bibinfo{year}{2016}, pp. \bibinfo{pages}{400--418}.
%Type = Misc
\bibitem[{Radford et~al.(2022)Radford, Kim, Xu, Brockman, McLeavey, and Sutskever}]{radford2022}
\bibinfo{author}{A.~Radford}, \bibinfo{author}{J.~W. Kim}, \bibinfo{author}{T.~Xu}, \bibinfo{author}{G.~Brockman}, \bibinfo{author}{C.~McLeavey}, \bibinfo{author}{I.~Sutskever}, \bibinfo{title}{Robust speech recognition via large-scale weak supervision}, \bibinfo{year}{2022}. \URLprefix \url{https://arxiv.org/abs/2212.04356}. \href{http://arxiv.org/abs/2212.04356}{{\tt arXiv:2212.04356}}.
%Type = Article
\bibitem[{Honnibal et~al.(2020)Honnibal, Montani, Van~Landeghem, Boyd et~al.}]{honnibal2020spacy}
\bibinfo{author}{M.~Honnibal}, \bibinfo{author}{I.~Montani}, \bibinfo{author}{S.~Van~Landeghem}, \bibinfo{author}{A.~Boyd}, et~al.,
\newblock \bibinfo{title}{spacy: Industrial-strength natural language processing in python}  (\bibinfo{year}{2020}).
%Type = Inproceedings
\bibitem[{McAuliffe et~al.(2017)McAuliffe, Socolof, Mihuc, Wagner, and Sonderegger}]{mcauliffe2017montreal}
\bibinfo{author}{M.~McAuliffe}, \bibinfo{author}{M.~Socolof}, \bibinfo{author}{S.~Mihuc}, \bibinfo{author}{M.~Wagner}, \bibinfo{author}{M.~Sonderegger},
\newblock \bibinfo{title}{Montreal forced aligner: Trainable text-speech alignment using kaldi.},
\newblock in: \bibinfo{booktitle}{Interspeech}, volume \bibinfo{volume}{2017}, \bibinfo{year}{2017}, pp. \bibinfo{pages}{498--502}.
%Type = Article
\bibitem[{Bradski(2000)}]{bradski2000opencv}
\bibinfo{author}{G.~Bradski},
\newblock \bibinfo{title}{The opencv library.},
\newblock \bibinfo{journal}{Dr. Dobb's Journal: Software Tools for the Professional Programmer} \bibinfo{volume}{25} (\bibinfo{year}{2000}) \bibinfo{pages}{120--123}.
%Type = Misc
\bibitem[{Lugaresi et~al.(2019)Lugaresi, Tang, Nash, McClanahan, Uboweja, Hays, Zhang, Chang, Yong, Lee, Chang, Hua, Georg, and Grundmann}]{lugaresi2019}
\bibinfo{author}{C.~Lugaresi}, \bibinfo{author}{J.~Tang}, \bibinfo{author}{H.~Nash}, \bibinfo{author}{C.~McClanahan}, \bibinfo{author}{E.~Uboweja}, \bibinfo{author}{M.~Hays}, \bibinfo{author}{F.~Zhang}, \bibinfo{author}{C.-L. Chang}, \bibinfo{author}{M.~G. Yong}, \bibinfo{author}{J.~Lee}, \bibinfo{author}{W.-T. Chang}, \bibinfo{author}{W.~Hua}, \bibinfo{author}{M.~Georg}, \bibinfo{author}{M.~Grundmann}, \bibinfo{title}{Mediapipe: A framework for building perception pipelines}, \bibinfo{year}{2019}. \URLprefix \url{https://arxiv.org/abs/1906.08172}. \href{http://arxiv.org/abs/1906.08172}{{\tt arXiv:1906.08172}}.
%Type = Article
\bibitem[{Serengil and Ozpinar(2024)}]{serengil2024lightface}
\bibinfo{author}{S.~Serengil}, \bibinfo{author}{A.~Ozpinar},
\newblock \bibinfo{title}{A benchmark of facial recognition pipelines and co-usability performances of modules},
\newblock \bibinfo{journal}{Journal of Information Technologies} \bibinfo{volume}{17} (\bibinfo{year}{2024}) \bibinfo{pages}{95--107}. \URLprefix \url{https://dergipark.org.tr/en/pub/gazibtd/issue/84331/1399077}. \DOIprefix\doi{10.17671/gazibtd.1399077}.
%Type = Misc
\bibitem[{Kasi(2002)}]{kasi2002yet}
\bibinfo{author}{K.~Kasi}, \bibinfo{title}{Yet another algorithm for pitch tracking (yaapt)}, \bibinfo{year}{2002}.
%Type = Inproceedings
\bibitem[{Eyben et~al.(2010)Eyben, W{\"o}llmer, and Schuller}]{eyben2010opensmile}
\bibinfo{author}{F.~Eyben}, \bibinfo{author}{M.~W{\"o}llmer}, \bibinfo{author}{B.~Schuller},
\newblock \bibinfo{title}{Opensmile: the munich versatile and fast open-source audio feature extractor},
\newblock in: \bibinfo{booktitle}{Proceedings of the 18th ACM international conference on Multimedia}, \bibinfo{year}{2010}, pp. \bibinfo{pages}{1459--1462}.
%Type = Article
\bibitem[{Sparck~Jones(1972)}]{sparck1972statistical}
\bibinfo{author}{K.~Sparck~Jones},
\newblock \bibinfo{title}{A statistical interpretation of term specificity and its application in retrieval},
\newblock \bibinfo{journal}{Journal of documentation} \bibinfo{volume}{28} (\bibinfo{year}{1972}) \bibinfo{pages}{11--21}.
%Type = Misc
\bibitem[{Zhao et~al.(2016)Zhao, Liu, and Xu}]{zhao2016sentiment}
\bibinfo{author}{J.~Zhao}, \bibinfo{author}{K.~Liu}, \bibinfo{author}{L.~Xu}, \bibinfo{title}{Sentiment analysis: Mining opinions, sentiments, and emotions}, \bibinfo{year}{2016}.
%Type = Misc
\bibitem[{Argaw et~al.(2024)Argaw, Yoon, Heilbron, Deilamsalehy, Bui, Wang, Dernoncourt, and Chung}]{argaw2024scaling}
\bibinfo{author}{D.~M. Argaw}, \bibinfo{author}{S.~Yoon}, \bibinfo{author}{F.~C. Heilbron}, \bibinfo{author}{H.~Deilamsalehy}, \bibinfo{author}{T.~Bui}, \bibinfo{author}{Z.~Wang}, \bibinfo{author}{F.~Dernoncourt}, \bibinfo{author}{J.~S. Chung}, \bibinfo{title}{Scaling up video summarization pretraining with large language models}, \bibinfo{year}{2024}. \URLprefix \url{https://arxiv.org/abs/2404.03398}. \href{http://arxiv.org/abs/2404.03398}{{\tt arXiv:2404.03398}}.
%Type = Inproceedings
\bibitem[{Narasimhan et~al.(2022)Narasimhan, Nagrani, Sun, Rubinstein, Darrell, Rohrbach, and Schmid}]{narasimhan2022tl}
\bibinfo{author}{M.~Narasimhan}, \bibinfo{author}{A.~Nagrani}, \bibinfo{author}{C.~Sun}, \bibinfo{author}{M.~Rubinstein}, \bibinfo{author}{T.~Darrell}, \bibinfo{author}{A.~Rohrbach}, \bibinfo{author}{C.~Schmid},
\newblock \bibinfo{title}{Tl; dw? summarizing instructional videos with task relevance and cross-modal saliency},
\newblock in: \bibinfo{booktitle}{European Conference on Computer Vision}, \bibinfo{organization}{Springer}, \bibinfo{year}{2022}, pp. \bibinfo{pages}{540--557}.

\end{thebibliography}

% \appendix


% The sources for the ceur-art style are available via
% \begin{itemize}
% \item \href{https://github.com/yamadharma/ceurart}{GitHub},
% % \item \href{https://www.overleaf.com/project/5e76702c4acae70001d3bc87}{Overleaf},
% \item
%   \href{https://www.overleaf.com/latex/templates/template-for-submissions-to-ceur-workshop-proceedings-ceur-ws-dot-org/pkfscdkgkhcq}{Overleaf
%     template}.
% \end{itemize}

\end{document}